\documentclass{llncs}
\usepackage{times}
\usepackage{url}
\usepackage{latexsym}

\usepackage{CJKutf8}

\usepackage{multirow}
\usepackage{amsmath}
\usepackage{threeparttable}
\usepackage{array} 
\usepackage[round, authoryear, sort]{natbib}
\usepackage{threeparttable}
\usepackage{makecell}
\usepackage{url}
\usepackage{bm}
\usepackage{amsmath}
\usepackage{amssymb}
\usepackage{graphicx}
\usepackage{subfigure}
\usepackage{caption}
\usepackage{mathrsfs}
\usepackage{csquotes}

\newlength\myequborder
\usepackage{csquotes}

\newenvironment{nospaceflalign*}
 {\setlength{\abovedisplayskip}{\myequborder}\setlength{\belowdisplayskip}{\myequborder}%
  \csname flalign*\endcsname}
 {\csname endflalign*\endcsname\ignorespacesafterend}

\title{An Event Network for Exploring Open Information}

\author{Yanping Chen\\
  {\tt ypench@gmail.com}
  \\}

\begin{document}
\begin{CJK}{UTF8}{gbsn}

\maketitle
\begin{abstract}
In this paper, an event network is presented for exploring open information, where linguistic units about an event are organized for analysing. The process is divided into three steps: document event detection, event network construction and event network analysis. First, by implementing event detection or tracking, documents are retrospectively (or on-line) organized into document events. Secondly, for each of the document event, linguistic units are extracted and combined into event networks. Thirdly, various analytic methods are proposed for event network analysis. In our application methodologies are presented for exploring open information.
\end{abstract}


\section{Introduction}
\label{sec:introduction}

Exploring information in open fields is a crucial and challenging task for human beings in the age characterized by flooding of information. Gigantic volumes of data spring up all over the world. This information is availably for us by connecting into the Internet. Events (e.g. the Ebola virus, the Islamic state) outbroken far away from us may affect our ordinary life. We no longer turn a blind eye to world events, that we are getting involved. We not only consume information, but also produce it. Social media (e.g., Twitter) makes us easily and freely express opinions. Millions of comments spread over the Internet, affecting the trend of public opinions. Accessing this information is beneficial for our decision-makings. Commonly, Information Retrieval (IR) and Information Extraction (IE) provide helpful solutions for us to handle the surge of information.

When exploring information, event oriented techniques provide effective approaches to understand who, where, when and what happened. Output of event detection in IR is series (or clusters) of documents, referred as {\em document events} in this paper. The problem with IR systems is that, they only retrieve subject-documents, but no content of documents is identified. Users are required to skim through returned documents. Events of IE are templates with slots to be filled, referred to as {\em template events}. Template event recognition extracts structured data from semi-structured or unstructured data. It is expected that the result is used to populate a knowledge base directly. The main problem for template event recognition is that it suffers from poor performance, especially in an open field, where heterogeneous resources, noise and fragmental data are processed. 

In open fields, there are knowledge bases automatically constructed to support information exploring. Many of them are constructed from semi-structured database (e.g. Wikipedia, WordNet) or under human supervision (collaboratively). Therefore, a better consistency is expected. They are widely used as external knowledge for semi-supervised methods. Because semi-structured data or human labor are required, when a news is broken in open fields automatically handling this information in real time is difficult. Furthermore, many systems automatically organize extracted linguistic units into a graph-based representation. These systems generally lead to a complex network with thousands of nodes or edges. Rare analysis was conducted to show underlying structures of events. Generally, output generated by IE systems is error-prone, redundant and incompatible information makes it contradictory. In this paper, analyses of event network are emphasized. Contributions of this paper include,
\begin{enumerate}
\item An event network framework is proposed, which provides an event oriented information exploring and supports various analyses.
\item Two novel analytic methods (PLT analysis and action analysis) are discussed for event oriented information exploring.
\end{enumerate}

The rest of this paper is organized as follows. Section \ref{sec:graph-based-representation} introduces related work. Motivations and definitions about the event network are presented in Section \ref{sec:motivation}. Section \ref{sec:event-network-construction} discusses our method to construct event networks. Section \ref{sec:application} demonstrates applications of event network. Section \ref{sec:conclusion} gives conclusions.

\section{Related Work}
\label{sec:graph-based-representation}

Organizing linguistic units as networks or graphs shows an increased interest in NLP. These representations can be roughly divided into three paradigms: logic based semantic network (e.g. Ontology), scalable knowledge database (e.g. Freebase) and semantic network constructed from open information extraction.

Logic based semantic network refers to networks constructed with human labor. They mainly focus on a closed domain. For example, {\em conceptual graph} represents logic as a graph representation \citep{sowa1984conceptual}. These networks support logic operators, can map questions and assertions from natural language to a relational database. Logic based semantic networks are constructed by domain experts and used as domain-specific ontologies (e.g., WordNet, Cyc, etc.). Commonly, they support inferences developed in knowledge representation. In this paradigm, conflicting and contradictory are not allowed. Therefore, in open fields it is difficult to apply.

In the second paradigm, logic based semantic network is extended into open fields. Large knowledge databases such as Yago \citep{suchanek2007yago}, Freebase \citep{bollacker2008freebase} are constructed, representing large knowledge in formalized forms. These representations merge diverse and heterogeneous data with high scalability, providing a unified framework for organizing information. Instead of rigid definitions, these networks have no canonical view of data. They use a loose representation to support scalability and extensibility. Many of them are constructed by merging ontologies (e.g. WordNet, OpenCyc) or extracted from semi-structured database (e.g. Wikipedia). To construct logic based semantic network, commonly, direct or indirect human interventions are involved (e.g. collaborative methods or searching logs).

Instead of aiming at semi-structured data, the third paradigm explores information in an open and dynamic field (mainly focusing on unstructured data). In this paradigm, weak supervision \citep{mintz2009distant, xuexploring} and bootstrapping methods \citep{kozareva2010learning, mcintosh2011relation, weld2009using, agichtein2000snowball} are employed. Scalable knowledge databases are used to guide the process, e.g., T{\scriptsize EXT}R{\scriptsize UNNER} \citep{banko2007open}, K{\scriptsize NOW}I{\scriptsize T}A{\scriptsize LL} \citep{etzioni2005unsupervised, etzioni2011open}, WOE \citep{hoffmann2010learning} and StatSnowBall \citep{zhu2009statsnowball}. Commonly, in these systems, nodes are named entities and edges are relations between them. All extracted results are combined into a large network. It often generate a network containing more than thousands of nodes and edges.

The notion of event is widely used for exploring information. \citet{piskorski2011online} presents an on-line news event extraction system. Each event is defined as a frame with slots filled by information extracted from clustered documents, where the pattern matching method is used. \citet{ramakrishnan2014beating} proposes an EMBERS system encoding events as frames. It is used to forecast \enquote{civil unrest} events in open fields. TwiCal extracts open-domain events from Twitter \citep{ritter2012open}, where events are identified by named entities. In the same data set, ET represents events as clusters of keywords \citep{parikh2013events}. \citet{kuzey2014fresh} uses events itself as nodes of network. They cluster documents into a hierarchical representation, where notes are events and edges link the same event in chronological order. \citet{angel2012dense} constructs an entity network from social media by the {\em streaming edge weight} method. They mine dense subgraphs of network for identifying realtime stories. \citet{das2011dynamic} provides an event discovery method based on entity dynamic relation graphs, which are constructed by co-occurrences of entities constrained in documents. 


\section{Motivation and Definition}
\label{sec:motivation}

Methodologies to organize linguistic units into networks or graphs show an increased interest in NLP. They provide novel solutions for many NLP tasks and support human oriented information exploring. Representing linguistic units as a graph enables topological analyses developed in fields such as: social network and complex network.

In open fields, these representations are mainly constructed by techniques developed under information extraction or text understanding. Information extraction aims at extracting linguistic units with concrete concepts or functions. It is seen as a trade-off between information retrieval and text understanding, where text understanding tries to capture all information in a document. Text understanding may lead to worsen performance caused by applied techniques \citep{hobbs2010information}. On the other hand, information extraction extracts targeted units and ignores uninterested.


Due to extracting challenges in open fields, instead of extracting information in a monolithic process, we divide the task into three steps: document event detection, event network construction and event network analysis. In the first step, by implementing document event detection and tracking, documents are organized into document events. Most of irrelevant or uninterested information is filtered. Then, in the second step, IE techniques are employed to extract linguistic units and organized into event networks. Techniques with higher performance are highlighted. For example, instead of extracting named entities as nodes, entity mentions are used. Where coreference resolution are required to group entity mentions into named entities, which is error-prone. In the last step, because topological information is available, structural information between linguistic units can be used to modify the network quality. Then, network or graph based analytic methods can be used, and it is convenient for visualization.

In this domain, many systems combine extracted results into a complex network, which not doing much help for analysing information. Furthermore, redundant and incompatible information makes it contradictory and misunderstanding. In our application, we emphasize methodologies conducted for event network analysis. Advantages of the event network include: {first}, after document event detection, information extraction in each document event can be independently implemented. Therefore, the effect of noise and heterogeneous data on information extraction can be reduced. {\ Secondly}, crossing document information enables discovery of potential relations between documents. {Thirdly}, event network provides a structured data representation for exploring open information, topological methods, e.g., social network or complex network, can be introduced for event network analysis. 


For convenience to discuss event network analysis, we define {\em nodes} and {\em edges} of event network as {\em frames} with {\em slots}. These information is also expected to support human oriented information exploring.

Let $\mathscr{D}\text{=}\{d_1, \cdots, d_L\}$ be a document set, $d_i$ denotes a document. A document event $\mathscr{E}_k$ is a subset of $\mathscr{D}$. For all $d_i, d_j \in \mathscr{E}_k$, similarity function $Similarity(d_i, d_j)$ satisfy a predefined condition (e.g. a threshold). All document events in $\mathscr{D}$ are denoted as $\mathscr{E}\text{=}\{\mathscr{E}_1, \cdots, \mathscr{E}_K\}$. The constraint that $\mathscr{E}$ is a partition of $\mathscr{D}$ is not necessary, because some documents in $\mathscr{D}$ can be filtered, or {\em fuzzy partitioning techniques} can be used, which enable a document belonging to more than one document event.

An event network on document event $\mathscr{E}_k$ is represented as a graph $\mathscr{N}_k\text{=}\{V_k, E_k\}$, where $V_k\text{=}\{v_{k1},\cdots,v_{kN}\}$ and $E_k\text{=}\{e_{k1},\cdots,e_{kM}\}$ are  vertex set and edge set. Both vertices and edges are frames defined as follows.
\begin{nospaceflalign*}
&\qquad\quad {vertex} := \{key, name, type, weight, info\}  & \nonumber\\
&\qquad\quad {edge} := \{type, v\text{-}1, v\text{-}2, weight, info \} & \nonumber
\end{nospaceflalign*}
where {\em vertex} frame defines nodes of event network. Slot \enquote{$name$} refers to entity mentions occurred in a document event. Each vertex is identified by an integer value \enquote{$key$}. Slot \enquote{$type$} represents categories of vertices (e.g. {\em Person}, {\em Organization} and {\em Location}). Slot \enquote{$weight$} is the likelihood of \enquote{$name$} to be \enquote{$type$}. Traditionally, this value is given by a classifier when extracting this frame. Depending on real applications, \enquote{$weight$} can be used to filter an event network. An {\em edge} frame denotes a relation between two vertexes. Slots \enquote{$v\text{-}1$} and \enquote{$v\text{-}2$} are $keys$ of vertices in an edge, used to identify vertices linked by edges. Edge types are referred by \enquote{$type$} (e.g. Part-whole, Personal-Social). In both frames, slot \enquote{$info$} contains information about the frames, where entity mentions or entity relations occurred, e.g., sentences, documents or timestamps. These information support event network analyses (e.g. coreference resolution, statistical relational learning or manually exploring). If they are empty, these values are $null$.



\section{Implementation}
\label{sec:event-network-construction}
This section discusses our method to construct event networks, which are used to show methodologies discussed in Section \ref{sec:application}.

\subsection{Data Sets}
We use the ACE 2005 Chinese corpus. It contains 633 documents annotated with 15,264 {entities} and 33,932 {entity mentions}\footnote{An {\em entity mention} is a reference to an entity.}. 7 entity types (e.g. {\em person}, {\em organization}, etc.) and 44 entity subtypes are defined. The corpus also annotated with 6 major relation types and 18 relation subtypes. Each relation instance has two named entities as arguments. There are 9,244 relation mentions are collected as positive instances.


The ACE 2005 Chinese corpus is used to train named entity and relation classifiers. In order to show our method in an open field, we also use the Chinese Gigaword Fifth Edition corpus. The Peoples Daily source is used, which contain 145,001  newswire texts covering the period from November 2006 through December 2010.

\subsection{Events Detection}
The purpose of document event detection is to cluster documents into events. We use LDA toolkit provided by \citet{phan2007gibbslda++} to implement this task. In LDA model, a corpus is first represented as a matrix, where each column refers to a document vector, and each row represents distribution of a term in documents. Then LDA maps documents from a term space into a topic space. Topics are hidden variables.

Because we focus on newswire texts, where short texts are commonly used. We use Omni-word feature proposed by \citet{chen-zheng-zhang:2014:P14-1}, which takes every potential word as terms of documents. It is a subset of n-Gram feature. In the pretreatment process, we remove high and low frequency words\footnote{The ratio is 5\% for each.} in an employed lexicon. Words with frequencies lower than 10 are also removed. To train an LDA model, hyper-parameters are required. The topic number is set as 25. Other parameters use default settings.

The toolkit generates several outputs. The word-topic distributions are more favourable to us, which give distributions of terms in a topic space. We use topics as centroids of document clusters in a term space. When clustering documents, a documents belonging to an event is judged by the nearest {\em Euler Distance} of the document and centroids. The top 100 most likely words per each topic are used to represent an event.


It is recognized that documents discussing the same event tend to be temporal proximity, and a time gap between bursts of similar documents may indicate different events \citep{yang1999learning}. Therefore, timestamps are used to partition the newswire texts. In our experiment, the time step is set as 5 months. Then, the Chinese Gigaword corpus is divided into 10 parts. Each part contains 5 months newswire texts. Because hierarchical representation can give a multi-granularity review when exploring open information and reduce the travel cost. In each time step, instead of using retrospective methods to give a flat partition of documents, we organize them into a hierarchical representation. Documents of each time step are clustered into 25 events by the LDA toolkit. Each event is further clustered into sub-events by the same approach. If an event contains documents less than ten, the process to find its sub-events is skipped. Therefore, in each time step, 25 events and at most $25 \times 25$ sub-events are detected.

\subsection{Named Entity Recognizing}
In this step, it is free to use any named entity recognition methods. In our application, we use a Boundary Assembling (BA) method  to implement the named entity recognition task. The notion of BA method is that, instead of recognizing entity mentions in a unitary style, it first detects boundaries of entity mention, then assembles detected boundaries into entity mention candidates. Each candidate is further assessed by a classifier.

In our work, we recognize three types of named entity: \enquote{PER} (Person),  \enquote{LOC} (Location) and  \enquote{ORG} (Organization). In order to filter noise, recognized named entities with Chinese characters less than two and more than six are discarded.

\subsection{Relation Recognizing}
To recognize relations between named entities, we adopt the method proposed in \citet{chen-zheng-zhang:2014:P14-1}, where an Omni-word feature and a soft constraint method is proposed for Chinese relation extraction. The Omni-word feature uses { every potential word in a relation mention} as lexical features. Then for each employed atomic feature, an appropriate constraint condition is selected to combine them with additional information to maximize the classification determination.

With our employed three entity types, five relation types annotated in the ACE corpus are recognized: \enquote{PER-SOC}, \enquote{GEN-AFF}, \enquote{ORG-AFF}, \enquote{PART-WHOLE} and \enquote{PHYS}. Sentences with more than ten entities are ignored, because extracting relations in a long sentence is error-prone.

\subsection{Merging and Visualizing}
As approaches discussed above, the result about recognized document events, named entities and relations are listed in Table \ref{tab:information-of-demonstrations}.

\begin{table}[h]
\begin{center}
\caption{Information of Results}
\label{tab:information-of-demonstrations}
\begin{tabular}{p{0.6cm}<{\centering}|r|c|r|r||p{0.6cm}<{\centering}|r|c|r|r}
\hline
{\bf Step} 	&\multicolumn{1}{c|}{\bf Doc.} &\multicolumn{1}{c|}{\bf Event} &\multicolumn{1}{c|}{\bf Entity} & \multicolumn{1}{c||}{\bf Relation}&{\bf Step} 	&\multicolumn{1}{c|}{\bf Doc.} &\multicolumn{1}{c|}{\bf Event} &\multicolumn{1}{c|}{\bf Entity} & \multicolumn{1}{c}{\bf Relation}\\\hline
{0} & {14,814} & {642}& {1,123,506} & {309,847}&{5} & {19,706} & {643}& {1,389,668} & {290,594}\\
{1} & {11,734} & {642}& {897,489} & {258,817}&{6}	& {12,014} & {645}& {927,977} & {265,439}\\
{2}	& {17,678} & {644}& {1,104,517} & {257,291}&{7} & {9,100}  & {641}& {705,805} & {254,133}\\
{3}	& {18,213} & {646}& {1,305,191} & {277,721}&{8} & {9,326}  & {635}& {726,123} & {230,535}\\
{4} & {23,433} & {644}& {1,704,294} & {289,166}&{9}	& {8,983}  & {640}& {674,457} & {243,210}\\
\hline
\end{tabular}
\end{center}
\end{table}

In the follows, in order to conduct event network analysis, we use the igraph toolkit provide by \citet{csardi2006igraph}, representing extracted entity mentions and relations as a graph. A network analysis toolkit (Pajek) provided by \citet{batagelj1998pajek} is used for visualization.

\section{Application}
\label{sec:application}

In our work, we emphasize analyses of event networks. After event networks were constructed, techniques such as social network, complex network can be employed to analyse event networks. For example, setting a person name as a central entity, we can navigate entities around it. Filtering irrelevant information, we can show character relationships in an event. Using the \enquote{PART-WHOLE} relations, multi-granularity visualization can be supported. Because event networks have a graph representation, topological information is available. Therefore, various approaches (e.g., statistical relational learning) can be used to improve the network quality. Furthermore, event networks support human oriented information exploring. When human exploring open information, manual interventions can be used to modify the quality of event networks.

\begin{figure}[h]
	\centering
	\includegraphics[width=6cm]{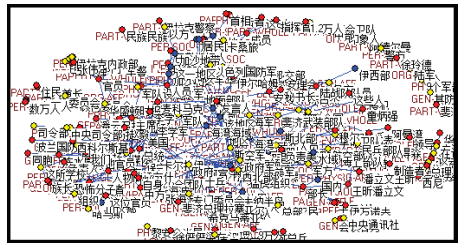}
	\caption{Event Network}
	\label{fig:event_network}
\end{figure}

In this section, we choose a document event in time step 0 as an example, which contain 1,041 documents. There are 42,436 named entities and 6,272 relations occurred. The most likely words in it are \enquote{袭击,北约,发言人,冲突,防御, etc.} (Assault, NATO, Spokesman, Conflict, Defence, etc.). It indicates that the concern of this event is military affairs. Extracted named entities and relations are organized in Figure \ref{fig:event_network}, where there are 252 nodes and 571 edges are merged. Nodes in {\em Red}, {\em Yellow} and {\em Blue} colors represent {\em Person}, {\em Organization} and {\em Location} respectively. Each edge is labelled by the relation type.


To explore open information, many systems dynamically organize linguistic units into a complex network. Heterogeneous resources and unreliable information make the network chaotic and misunderstanding. As Figure \ref{fig:event_network} showing, a complex network makes it difficult to understand. In the following, based on the event network, we give four methodologies to explore open information: Information Filtering, PLT Analysis, Action Analysis and Social Network Analysis.

\subsection{Information Filtering}

\begin{figure}[htbp]
\centering
\subfigure[Information Filtering]{
\begin{minipage}[h]{0.48\textwidth}
\includegraphics[width=6cm]{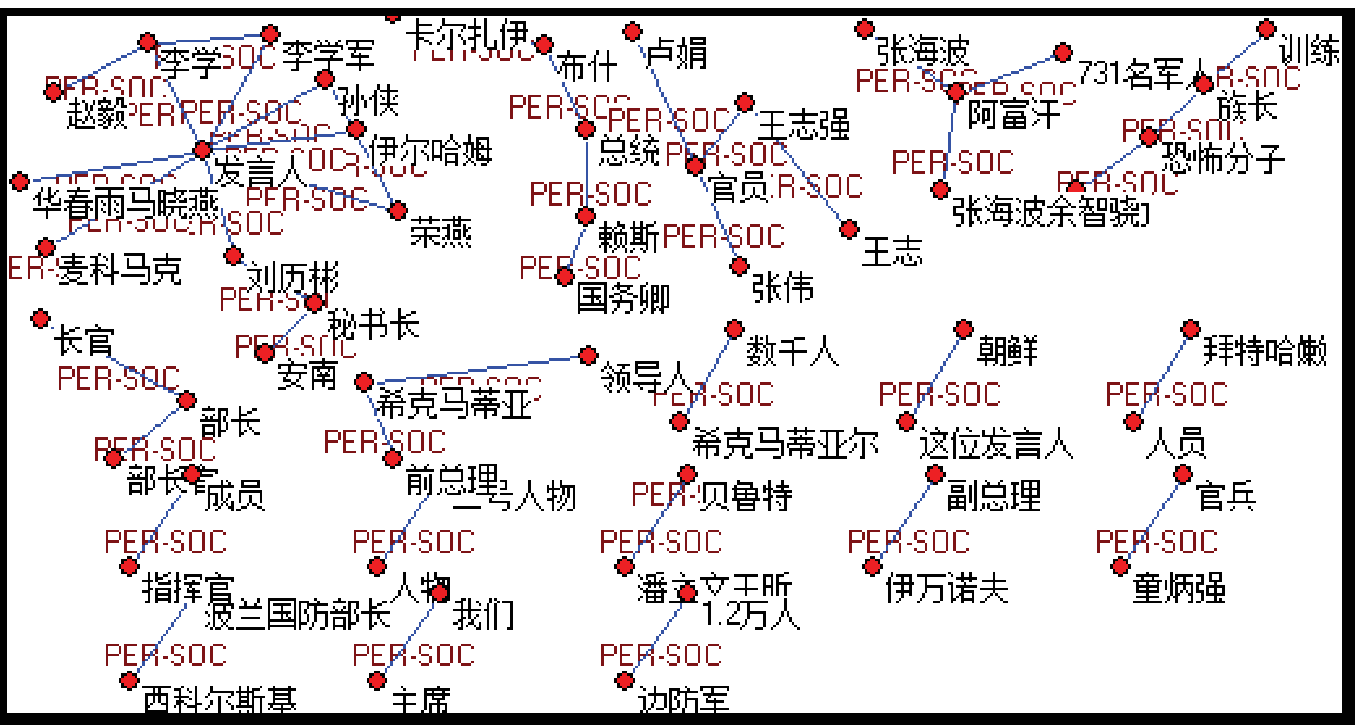}
\centering
\label{fig:information_filtering}
\end{minipage}
}
\subfigure[PLA Analysis]{
\begin{minipage}[h]{0.48\textwidth}
\includegraphics[width=6cm]{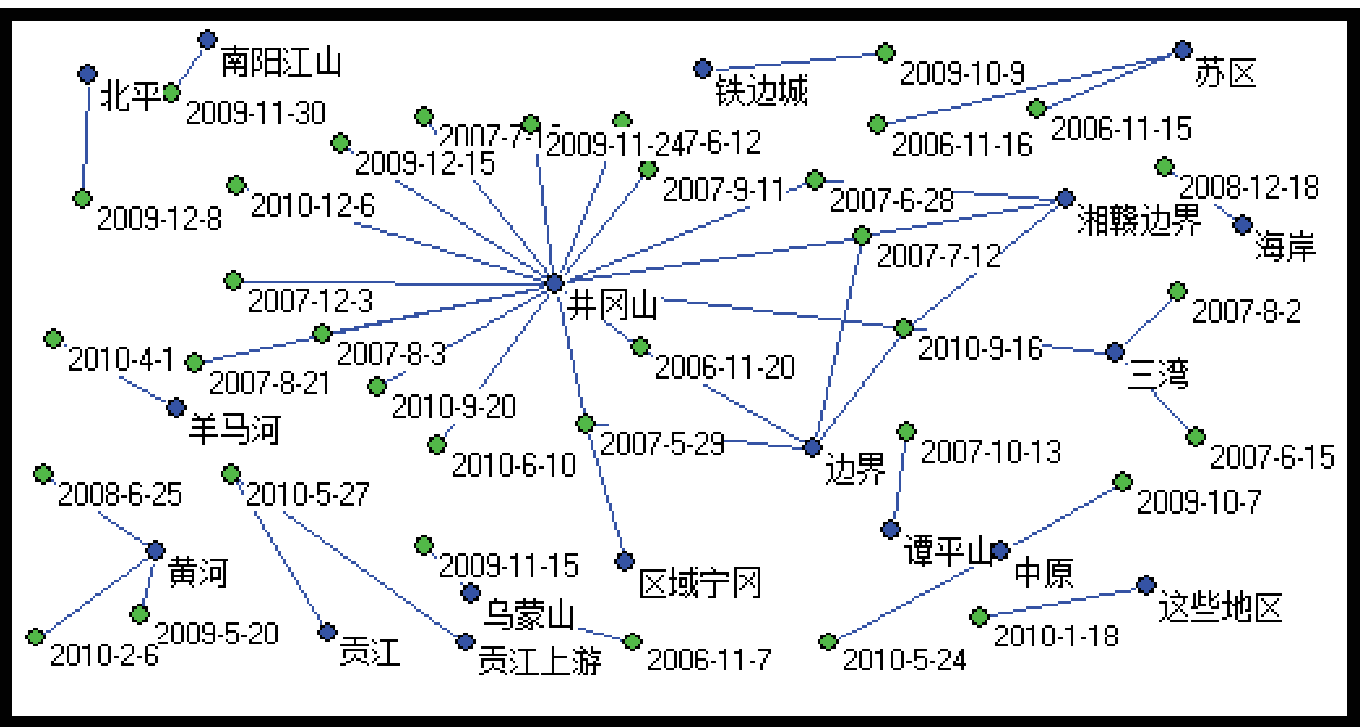}
\centering
\label{fig:pla_analysis}
\end{minipage}
}
\end{figure}

The simplest way to analyse event network is to filter information that is irrelevant or uninterested. In Figure \ref{fig:information_filtering}, only person names and \enquote{PER-SOC} relations are remained to show character relationships in an event network.

This example can be formalized as: Let $\mathscr{N}$ be an event network. The filtered event network $\mathscr{N}'\text{=}\{V', E'\}$ is a subgraph of $\mathscr{N}$, such that $\mathscr{N}' \subset \mathscr{N}$. And $\mathscr{N}'$ satisfies $\forall v \in V'(v.type = \text{PER}) \wedge \forall e \in E' (e.type = \text{PER-SOC})$.

Using information contained in $vertex$ frames and $edge$ frames, information filtering can provide effective approaches for exploring open information. For example, in $vertex$ frames and $edge$ frames, we may require that the value in $weight$ slots is greater than a predefined threshold. Utilizing information in $info$ slots, we can collect named entities occurred in specified periods or areas. For a central figure, we can see directly connected named entities and relations between them.

\subsection{PLT Analysis}
Person-Location-Time (PLT) analysis tries to find relations between persons and locations in a period of time. It can be used to track a person, find trajectories of targeted entities. $Person$, $Location$ are extracted by named entity recognition methods. While the $Time$ is different. Two kinds of $Time$ are distinguished in a document: {\em implicit temporal information} and {\em explicit temporal information}. Implicit temporal information is part of the document\textquoteright s content indicating the creation, development, termination of an event. It is also seen as a named entity type in some researches. Extracting this information needs information extraction or text understanding techniques. In many applications, it is ignored. Generally, in open fields, all documents have explicit temporal information, which includes the creation, modification and transmission timestamps of documents. They are meta-data spread with documents. In our paper, because we focus on newswire texts, where the explicit temporal information of documents is released together. Therefore, we use the explicit temporal information for PLT analysis. 

This process can be formalized by introducing an attribute $time$ in the $info$ slot. Let $\mathscr{N}$ be an event network, $times$ represent timestamps and $person$ is a person name. $\mathscr{N}'\text{=}\{V', E'\}$ is the result of PLT analysis based on $\mathscr{N}$, where $\forall e \in E' (e.type = \text{PHYS} \land (e.v\text{-}1 = person \lor e.v\text{-}2 = person))$. In other words, all relation type in $E'$ is \enquote{PHYS}, and take the same entity mention $person$ as an argument. Replacing all $person$ by corresponding $times$, we get a graph with nodes referred to timestamps and locations. An example is shown in Figure \ref{fig:pla_analysis}.


In this example, we track Mao Zedong (\enquote{毛泽东})\footnote{The leader of the Communist Party of Chinese.} in the whole Gigaword corpus, collect all recognized \enquote{PHYS} relation instances which have Mao Zedong as an argument. In the result, there are 142 \enquote{PHYS} relation mentions, which take Mao Zedong (or Chairman Mao) as arguments.  Then we replace Mao Zedong (or Chairman Mao) by the explicit temporal information of newswire texts. In Figure \ref{fig:pla_analysis}\footnote{Because the original graph is more complex (55 nodes and 142 edges), in this place, only part of it is given.}, nodes in green color are timestamps, and blue nodes are locations. Each green node means that Mao Zedong occurred with the connected locations at that time. 

\subsection{Action Analysis}

Recognizing an \enquote{event} under the ACE definition is difficult, where event triggers, participant roles, properties and attributes should be identified \citep{doddington2004automatic}. It received an ACE value score only 30\% in \citet{ahn2006stages}. In an open field, it will come to worse performance. In many researches, co-occurrence information (e.g. co-citation, co-word, co-link, etc.) between terms is used to explore and understand structures in the underlying document sets, e.g., \citet{leydesdorff2006co}. In our application, instead of the definition in ACE, we present the action analysis.

In action analysis, we focus on detecting whether or not a special action is mentioned in a sentence. Therefore, we conduct the \enquote{sentence classification} task, classing each sentence by a classifier trained on the ACE annotated event mentions. In our application, we monitor the \enquote{Conflict} ACE event type, which has two substype: Attack and Demonstrate \citep{doddington2004automatic}. The ACE corpus, which annotates 596 \enquote{Conflict} events, is employed for training and testing. We implement the 5-fold cross validation, and the P/R/F (Precision/Recall/F-score) measurement. F-score is computed by $(2 \times P \times R)/(P+R)$. In order to perform a two-class classification, we generate negative instances by segmenting the corpus into sentences, discarding annotated ACE event mentions, and filtering sentences without event triggers of \enquote{Conflict} ACE events. Then, 1,589 sentences are collected as negative instances. We only use Omni-words features in sentences for classification. The performance is shown in Row 1 of  Table \ref{tab:performance_of_action_analysis}, where Only the performance about \enquote{Conflict} is listed. 

\begin{table}[h]
\begin{center}
\caption{Performance of Action Analysis}
\label{tab:performance_of_action_analysis}
\begin{tabular}{p{0.8cm}<{\centering}|p{1.2cm}<{\centering}p{1.2cm}<{\centering}p{1.2cm}<{\centering}}
\hline
{\bf No.} 	& \bf Precision & \bf Recall & \bf F-core\\\hline
{1} & {82.19}& {82.88} & {82.53}\\
{2} & {97.65} & {41.94} & {58.68}\\
\hline
\end{tabular}
\end{center}
\end{table}

\begin{figure}[htbp]
\centering
\subfigure[Action Analysis]{
\begin{minipage}[h]{0.48\textwidth}
\includegraphics[width=6cm]{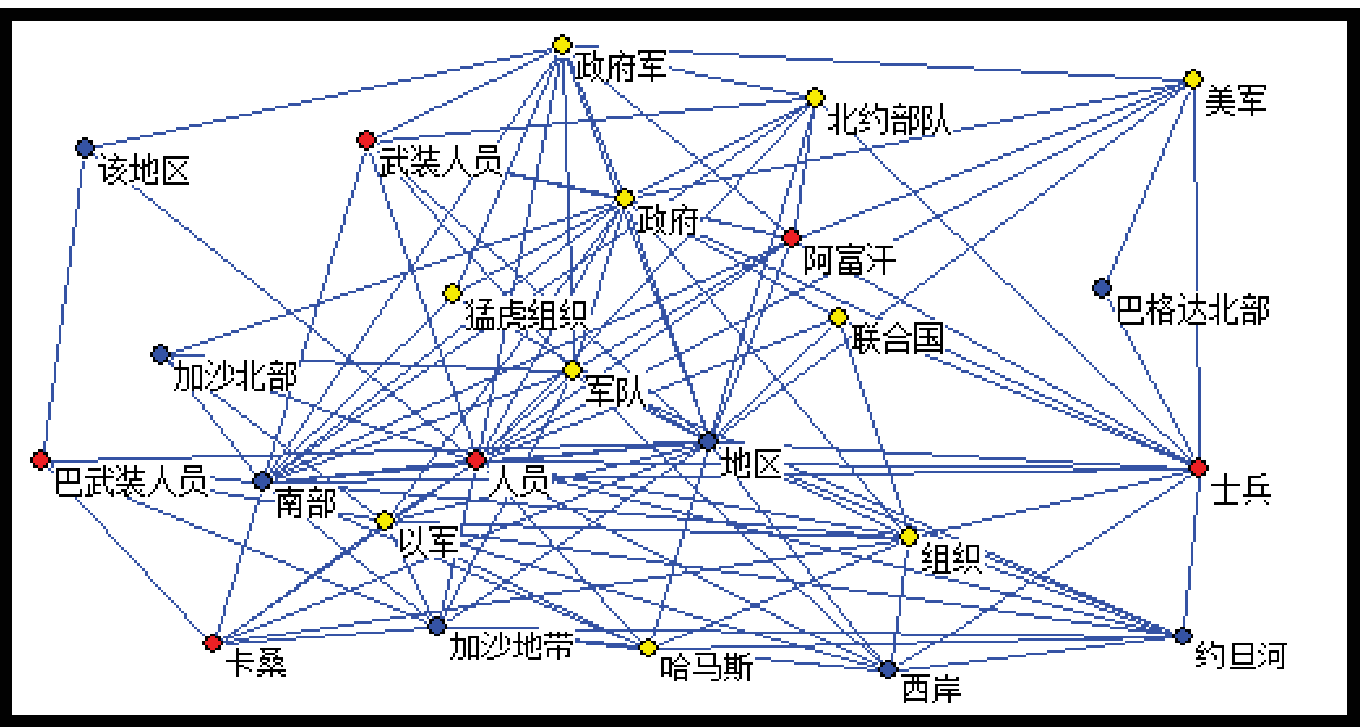}
\centering
\label{fig:action_analysis}
\end{minipage}
}
\subfigure[Social Network Analysis]{
\begin{minipage}[h]{0.48\textwidth}
\includegraphics[width=6cm]{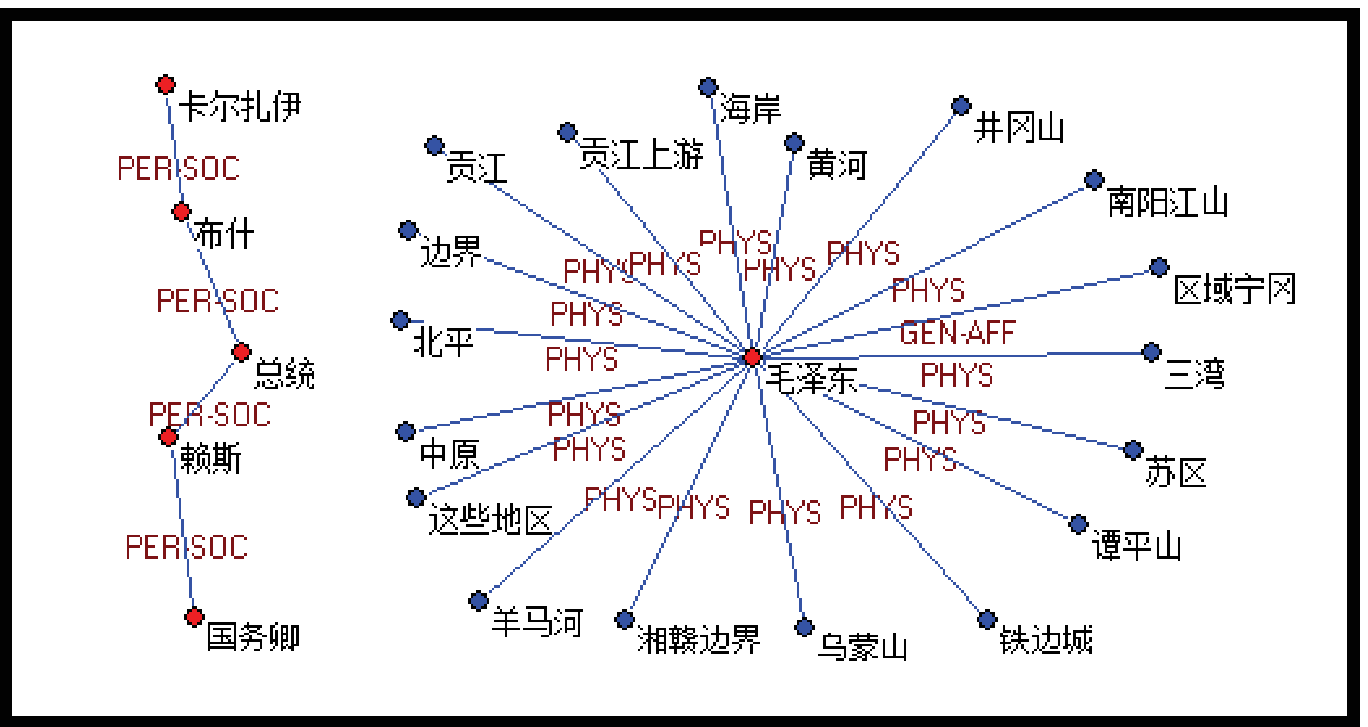}
\centering
\label{fig:social_analysis}
\end{minipage}
}
\end{figure}

In an open field with massive data, the precision is more emphasized. Therefore, we label an instance as a \enquote{Conflict} action only when the employed classifier (maximum entropy) output a predicted value equals 1 \footnote{The default value is 0.5 in two-class classification.}. The performance is shown in Row 2 of Table \ref{tab:performance_of_action_analysis}. We use this setting to train a classifier and predict every sentence in document events. Entity co-occurrences in each \enquote{Conflict} sentence are calculated. The result is shown in Figure \ref{fig:action_analysis}.


Figure \ref{fig:action_analysis} shows the result about the employed event. Edges in this example indicate co-occurrence relations between entities. In this event, there are 12,076 sentences containing at least two entities,where 836 sentences have the \enquote{Conflict} action with value 1 outputted by the classifier. Among them, 3,221 entities co-occurred. In order to make the result more comprehensible, edges with co-occurrence frequencies less than 12 are erased. Finally, a network with 25 entities is generated. In this example, entities (e.g., \enquote{哈马斯} (Hamas), \enquote{加沙北部} (the Gaza Strip), \enquote{阿富汗} (Afghanistan), \enquote{美军} (U.S. forces)) and the edges between them surely show meaningful information.

\subsection{Social Network Analysis}
Techniques (e.g., Short Path, Cohesive Subgroup, Center, etc.) proposed in social network mainly implemented on a network constructed by domain experts. A precise network is required to discover the underlying structure of social network. Because event networks are automatically extracted. They are error-prone. Therefore, for some of these techniques, it is difficult to get a reliable output. However, some results generated by social network also show meaningful information for us. In Figure \ref{fig:social_analysis}, an example is given.


Data in this example comes from results of {\em Information Filtering} and {\em PLT analysis}. The left of Figure \ref{fig:social_analysis} seeks a short path between \enquote{卡尔扎伊} (Hamid Karzai) and \enquote{国务卿} (the Secretary of State). They are connected by \enquote{PER-SOC} relations. On the right, \enquote{Mao Zedong} is set as the central figure to show directly collected locations, e.g., \enquote{井冈山} (Jinggangshan). 




\section{Conclusion and Future Work}
\label{sec:conclusion}

Event network is a framework for exploring open information. In this paper, based on the employed data set, we show applications of event network for information analyses. In future work, based on event network, more analyses can be developed to support exploring open information.


\bibliography{reference}

\end{CJK}
\end{document}